\documentclass{styles/svproc}
\usepackage{url}
\usepackage{cite}

\usepackage{amsmath,amssymb,amsfonts}
\usepackage{algorithmic}
\usepackage{graphicx}
\usepackage{textcomp}
\usepackage{xcolor}
\usepackage{color}
\usepackage{graphicx}
\usepackage{subfigure}
\usepackage{float}    
\usepackage{wrapfig}
\usepackage{color}
\usepackage{array}
\usepackage{graphicx}
\usepackage{multirow}

\def\MU#1{{\color{black} {{#1}}}}
\def\Rev#1{{\color{black} {{#1}}}}

\begin{document}

\mainmatter              
%
\title{Gaussian Process Distance Fields Obstacle and Ground Constraints for Safe Navigation
 }
\titlerunning{GPDF Constraints for Safe Navigation}
%

\author{Monisha Mushtary Uttsha \and Cedric Le Gentil \and Lan Wu \and Teresa Vidal-Calleja}

\authorrunning{M. Uttsha et al.} 


 \institute{UTS Robotics Institute, Faculty of Engineering
and IT, University of Technology Sydney, Sydney, NSW 2007, Australia\thanks{This work was supported by ARIA Research and the Australian Government via the Department of Industry, Science, and Resources CRC-P program (CRCPXI000007) and the Australian Research Council Discovery Project under Grant DP210101336.},\\
\email{monishamushtary.uttsha@student.uts.edu.au}}
\maketitle              

\begin{abstract}
Navigating cluttered environments is a challenging task for any mobile system. Existing approaches for ground-based mobile systems primarily focus on small wheeled robots, which face minimal constraints with overhanging obstacles and cannot manage steps or stairs, making the problem effectively 2D. However, navigation for legged robots (or even humans) has to consider an extra dimension. This paper proposes a tailored scene representation coupled with an advanced trajectory optimisation algorithm to enable safe navigation. Our 3D navigation approach is suitable for any ground-based mobile robot, whether wheeled or legged, as well as for human assistance. Given a 3D point cloud of the scene and the segmentation of the ground and non-ground points, we formulate two Gaussian Process distance fields to ensure a collision-free path and maintain distance to the ground constraints. Our method adeptly handles uneven terrain, steps, and overhanging objects through an innovative use of a quadtree structure, constructing a multi-resolution map of the free space and its connectivity graph based on a 2D projection of the relevant scene. Evaluations with both synthetic and real-world datasets demonstrate that this approach provides safe and smooth paths, accommodating a wide range of ground-based mobile systems.

\end{abstract}
\section{Introduction}

Safe navigation is one of the key requirements for robots operating in the real world. It has been a focus of robotics research for many years, with many algorithms developed to provide appropriate path plans given a representation of the environment. The difficulty of the task varies depending on the complexity of the environment and the adequacy of the representation. This is especially true in cluttered spaces, as it becomes increasingly difficult to plan a safe path that satisfies the embodiment constraints. Effective path planning for mobile robotic systems requires scene representations that consider the robot’s physical constraints and prioritise safety, especially when guiding impaired individuals. For example, visually impaired persons need safer paths than robots. With this motivation in mind, our work aims to develop a scene representation that facilitates the computation of safe paths for any ground-based system, including both wheeled and legged robots, as well as navigation assistance for humans.

Many scene representations for navigation have been proposed in the literature, such as occupancy maps~\cite{Octomap} and Euclidean Distance fields~\cite{Voxblox, wu2021faithful}. The essential feature of these representations is their capability to analyse free space, enabling planning algorithms to find collision-free paths. For occupancy mapping, the information is classified into three classes, observed free, observed occupied and unknown. This representation cannot explicitly tell how far the path is to a possible collision. Distance fields, on the other hand, have readily available information about the distance to the nearest surface for any point in the mapped environment. Thus, planning with distance fields allows more flexibility, as shown in \cite{ratliff2009chomp}. This can then be further augmented with the tree data structures to increase efficiency \cite{funk2021multi}.

Safety in path planning is not always guaranteed by commonly used planners such as A$^*$, or probabilistic roadmaps (PRM) \cite{prm_org}. Trajectory optimisation methods such as covariant Hamiltonian optimisation for motion planning (CHOMP)~\cite{Chomp}, on the other hand, allow safety constraints to be integrated. Recent works~\cite{wu2023log,men2020generalization} have integrated distance fields with such planners to generate collision-free paths in 2D for mobile robots~\cite{wu2023log} and in 3D for aerial vehicles~\cite{men2020generalization}. The aerial vehicles do not have ground constraints, and therefore, the extension from 2D to 3D is trivial given the appropriate dimension distance field. However, with ground constraints, planning given 3D distance fields requires further considerations, i.e., the mobile system must remain on the ground and account for overhanging objects, slopes or steps. 

This work presents a safe navigation approach that reasons on free space via continuous distance fields. Given a 3D point cloud of the environment, the segmentation of the ground points and the physical characteristics of the mobile system, we formulate two Gaussian Process distance fields (GPDFs) that feed a trajectory optimisation planner to produce collision-free paths that remain on the ground. Moreover, by projecting 3D points into 2D and exploiting the multiresolution aspects of a quadtree data structure and its connectivity graph, we are able to manage the free space efficiently without compromising safety. We propose a variation of CHOMP that not only accounts for collision avoidance but also for ground constraints, both encoded by different GPDFs. By using this novel dual-field approach, we ensure safety in all three dimensions while keeping the trajectory at a desired height above the ground. We validate our method using three publicly available real-world 3D datasets, two in their original forms and one augmented with two additional obstacles, to demonstrate the flexibility of our approach in regards to uneven surfaces, obstacles away from ground etc. Our findings demonstrate the method's suitability for both human navigation in intricate 3D environments and a variety of ground-based mobile robots.

\section{Related Work}

Distance field representations have become quite popular for mapping and planning in recent times. The emergence of efficient distance field representations such as Signed distance fields~ \cite{oleynikova2016signed}, truncated signed distance fields (TSDF) \cite{kinectfusion}, and Euclidean distance fields (EDF) \cite{Voxblox, han2019fiesta, wu2021faithful} have shown real-time applicability in motion planning. In particular, continuous distance fields using Gaussian processes have gained significant traction due to their ability to query distance at arbitrary resolutions and incorporate uncertainty. Works such as Log-GPIS \cite{wu2021faithful} and Gaussian Process Distance Field with a reverting function \cite{le2023accurate} have showcased accurate distance field estimation. 
These approaches can then be coupled with motion planners to achieve collision-free paths, as shown in \cite{wu2023log}. Similarly, discrete distance field approaches have shown the ability to be coupled with trajectory optimisation frameworks to produce safe navigation in 3D~\cite{men2020generalization}.

Motion planning is a well-studied field where classical graph search algorithms like A* are adapted for safe navigation and robot constraints. For instance, \cite{rioux2015humanoid} introduces Anytime Repairing A* (ARA), which initially finds sub-optimal paths and then optimises them according to the robot's kinematic constraints. Liu et al. \cite{teng2023motion} propose a graph search method combined with an ellipsoid model for quadrotor navigation through tight spaces. Marcucci et al. \cite{marcucci2024fast} create a weighted graph of safe spaces offline, then find the shortest path using a graph search method. Additionally, sampling-based approaches like RRT and PRM are popular for safe navigation, as demonstrated in \cite{muhammad2022simulation}, where these methods find obstacle-free paths for autonomous systems.


Even though such planners can provide collision-free paths, there is usually no explicit control over the distance maintained from obstacles. This can be addressed by considering graph nodes or samples a certain distance away, as in \cite{marcucci2024fast}, though this is not a generalised solution and may not always find a suitable path. Trajectory optimisation methods effectively tackle this issue. One of the most popular frameworks is CHOMP \cite{ratliff2009chomp}, which uses a distance field and its gradient to maintain a defined distance from obstacles. As an optimisation approach, it allows for designing a cost function that ensures smoothness while maintaining the desired distance. Zucker et al. \cite{zucker2013continuous} used CHOMP to plan and execute smoother trajectories for a NAO robot in a door-opening task. Men et al. \cite{men2020generalization} proposed a CHOMP variation for avoiding dynamic obstacles with a UAV, demonstrating the potential of trajectory optimisation to constrain trajectories for specific applications through appropriate cost function design. In this work, we propose a fit-for-purpose cost function considering ground constraints and obstacles within the robot's height to enable safe navigation and considering the size of the system. 

Coupling trajectory optimisation methods with multi-resolution data structures have shown efficient and accurate performance. 
He et al. \cite{he2013multigrid} proposed a variation of~\cite{Chomp} to ensure a fast run-time of the CHOMP algorithm constrained with the kinematic constraints of the robot. A drawback of this work, however, is that they assume the paths have already been validated for collision-free motion before being fed into the optimisation. Authors in \cite{han2019fiesta} introduced efficient grid maps and indexing data structures for their planning framework. Reijgwart et al. \cite{reijgwart2024waverider} developed an obstacle avoidance policy using a multiresolution volumetric map and Riemannian motion policies (RMP) \cite{ratliff2018riemannian} for micro aerial vehicle navigation in indoor spaces. Funk et al. \cite{funk2021multi} proposed using a multiresolution grid to map free space, enabling fast collision queries and planning.

Some of the discussed works focus on mobile robots prioritising optimality in terms of accuracy and smooth motion. However, adhering to safety and feasibility requirements is equally important, as in the work presented here. Other work, such as in \cite{barbosa2022towards}, propose smooth approximations of Euclidean distance fields, coupled with a control barrier function for collision avoidance, resulting in a small, risk-aware subset of obstacle-free space. Majd et al. \cite{majd2021safe} present an RRT planner augmented with a control barrier function (CBF) to ensure safe navigation in narrow corridors, tested on a simulated dataset. Planning for humans and prioritising safety in complex 3D environments remains an open challenge. Our work aims to find feasible, safe paths for ground-based mobile systems, including humans, in real-world cluttered environments.


\section{Preliminaries} \label{background}
\subsection{GP Distance Field with Reverting Function}
 Gaussian processes (GP) is a non-parametric, machine learning algorithm used for both classification and regression tasks. A GP describes the probability distribution over the functions $f(\mathbf{x}) $ that fit over the given dataset \(\mathbf{X} = [\mathbf{x}_1, \mathbf{x}_2, \ldots, \mathbf{x}_n]\) and its noisy measurements \( \mathbf{y} = [{y_1}, {y_2}, \ldots, {y_n}] \). The mean function is often taken as zero or constant, and its covariance matrix is obtained by a kernel function evaluated at all input pairs in its vanilla form \cite{rasmussen2003gaussian}. 
As in \cite{le2023accurate}, let us formulate a GP as ${o(\mathbf{x})\sim \mathcal{GP}(0,k_o(\mathbf{x},\mathbf{x}'))}$ to learn a continuous function that depicts the occupancy of the scene given a registered point cloud. 
This latent scalar field can be obtained given the observations $\mathbf{x}_i$, from the surface where ${y_i}=1$. 
Using the standard GP joint probability distribution formulation, the occupancy at any point $\mathbf{x}$ can be inferred via

\begin{equation}
\hat{o}(\textbf{x}) = \mathbf{k_{xX}} \left( \mathbf{K_{XX}} + \sigma_o^2 \mathbf{I} \right)^{-1} \mathbf{1},
\label{occupancy}
\end{equation}
where evaluating the $k_o(\mathbf{x},\mathbf{x}')$, gives the $\mathbf{K_{XX}}$ matrix that represents the covariance of the noisy surface observations and
$\mathbf{k_{xX}}$ is the covariance between the testing point and the observations. $\sigma_o^2$ is the variance of the observations' noise and $\mathbf{I}$ is an identity matrix with the same size as $\mathbf{K_{XX}}$. \\

Following the formulation in~\cite{le2023accurate}, if the kernel function is reversed, a reverting function $r$ can be formed, which maps occupancy to their corresponding distance values. Therefore, we query the occupancy through \eqref{occupancy} and recover analytically (subject to the chosen kernel) the reverted form to obtain the distance field as, 






\begin{equation}
    \hat{d}(\textbf{x}) = r\left( \hat{o}(\textbf{x}) \right) \quad \text{with} \quad r\left( k_o(\textbf{x}, \mathbf{x'}) \right) \triangleq \| \textbf{x}- \mathbf{x'} \|,
\end{equation}
 where the reverting function $r$ would vary depending on which kernel function is being used to define the occupancy field in \eqref{occupancy}.

 
 Following the principles of linear operators as detailed in \cite{sarkka_pde}, the gradient of the distance function obtained with reverting can be written as:
\begin{equation}
    \nabla_{x} \hat{d}(\textbf{x}) \approx \nabla_{x} r(\hat{o}(\textbf{x}))\approx\nabla_{o} r(\hat{o}(\textbf{x})) \nabla_{x} \hat{o}(\textbf{x}).
\end{equation}

Note that, the distance field can be formulated in both 2D and 3D, for points on the ground and for points on the obstacles as we will show later. Furthermore, in this work we use formulation of the unsigned Euclidean distance field.
\vspace{-3ex}
\subsection{Trajectory Optimisation Planner}

Trajectory optimisation methods refine an initial trajectory by iteratively minimising costs such as collision avoidance and constraint violations. CHOMP~\cite{Chomp} focuses on creating a smooth, collision-free path through non-linear optimisation on an estimated initial trajectory. Let \( \mathbf{x}^r(t):\mathbb{R} \rightarrow \mathbb{R}^D\) denote the robot's trajectory in D-dimensional space at time $t$.



The objective function, as introduced in \cite{ratliff2009chomp} is as follows,  
\begin{equation}
C[\mathbf{x}^r] \equiv \int_{0}^{T} c_s((\mathbf{x}^r(t))+ \lambda c(\mathbf{x}^r(t)) \| \mathbf{\dot{x}}^r(t) \| \, dt
\label{chomp_obj}\,,
\end{equation}
where the total cost $C[\mathbf{x}^r]$ is minimised over time $0$ to $T$. Obstacle avoidance is enforced by the collision penalty $c(\mathbf{x}^r(t))$. 
CHOMP takes in the distance value, $d(\mathbf{x}^r(t))$, to the closest surface, and thus calculates the collision cost based on:
\begin{equation}
c(\mathbf{x}^r(t)) =
\begin{cases}
-d(\mathbf{x}^r(t)) + \frac{1}{2} \epsilon, & \text{if } d(\mathbf{x}^r(t)) < 0 \\[10pt]
\frac{1}{2\epsilon} \left( d(\mathbf{x}^r(t)) - \epsilon \right)^2, & \text{if } 0 < d(\mathbf{x}^r(t)) \leq \epsilon \\[10pt]
0, & \text{otherwise.}
\end{cases}
\label{obs_cost}
\end{equation}
The term $c_s((\mathbf{x}^r(t))$  encourages smoothness with the regularisation of the velocity calculated as 
\begin{equation}
c_s((\mathbf{x}^r(t))=\frac{1}{2} \| \mathbf{\dot{x}}^r(t) \|^2 .
\end{equation}

It is minimised using covariant gradient descent to obtain the final trajectory. 

\section{Methodology} \label{method}
\begin{figure}
    \centering
        \includegraphics[scale=0.4]{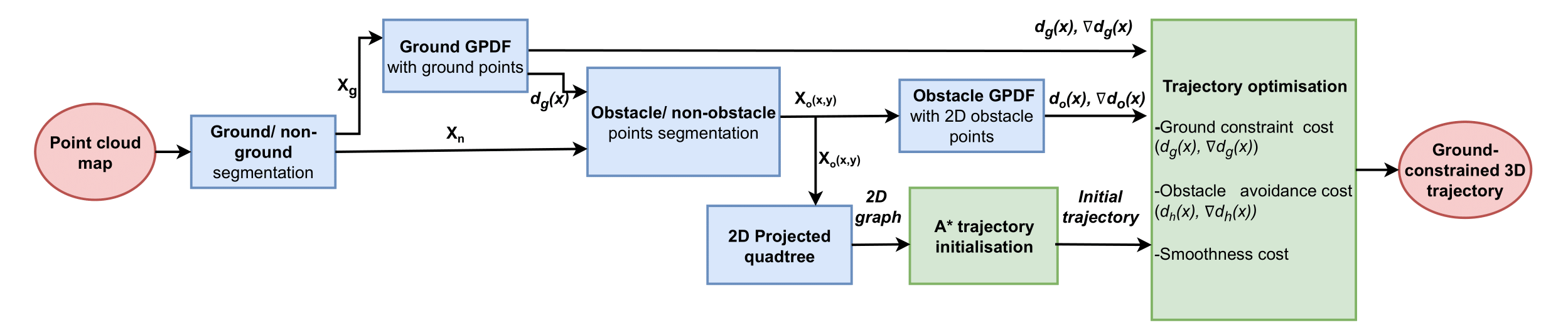}
     
    \caption{Block diagram of the proposed approach. All steps related to representation are colored in light blue. Steps related to planning are colored in light green.}
    \label{pipeline}
    \vspace{-4ex}
\end{figure}

In this section, we detail our proposed approach for safe navigation, which includes the scene representation and the optimisation-based planner. The overview of the proposed framework is shown in  Fig. \ref{pipeline}.

\subsection{Overview}
Let us consider a cylindrical, ground-based system with a height $h_r$ moving in a cluttered 3D environment. The aim of this work is to find a trajectory for the given system, such that it remains collision-free along its cylindrical shape, while preventing the system from lifting off the ground. 
To this end, we propose the use of a dual scene representation with two Gaussian Process distance fields that are coupled with a modified version of a trajectory optimisation method to produce safe and feasible paths.

As shown in Fig. \ref{pipeline}, the pipeline takes the 3D point cloud map of the scene and segments it into the ground $\mathbf{X}_g$ and non-ground $\mathbf{X}_n$ points sets. Given the ground points, we first build the \emph{Ground} GPDF $d_g$. The Ground GPDF can provide the height of any waypoint from the ground and it is particularly useful when the ground is not flat. Amongst the non-ground points, only the points within the system's maximum height, are of concern for obstacle avoidance. Thus, the non-ground points are further classified into obstacle points $\mathbf{X}_o$, using the Ground GPDF and the system's height. The obstacle points are then projected into a 2D plane to build the second GPDF, called \emph{Obstacle} GPDF. The key idea is that the Ground GPDF will be used to satisfy the ground constraints while the Obstacle GPDF facilitates obstacle avoidance within the system's height.  
\\
With the projected set of points, we further formulate a quadtree and its connectivity graph, which is used to create an initial trajectory approximation with A* graph search. We then use this to optimise and constrain the motion, using a modified CHOMP algorithm aided by the Ground GPDF and the Obstacle GPDF. 
We propose a modification of the objective function of CHOMP in \eqref{chomp_obj} and its gradient to account for the ground constraints along with smoothness and collision avoidance. Finally, using the covariant gradient descent optimisation, we obtain our desired trajectory that is collision-free, safe, smooth and feasible. 

\subsection{Scene Representation with Distance Fields}
\label{sec:rep}
We choose to represent the environment using Gaussian Process distance fields that enable models for ground and obstacle separately and that can be inferred at arbitrary points.

\begin{figure}
    \centering
    \subfigure[Point cloud map] {
        \includegraphics[width=30mm]{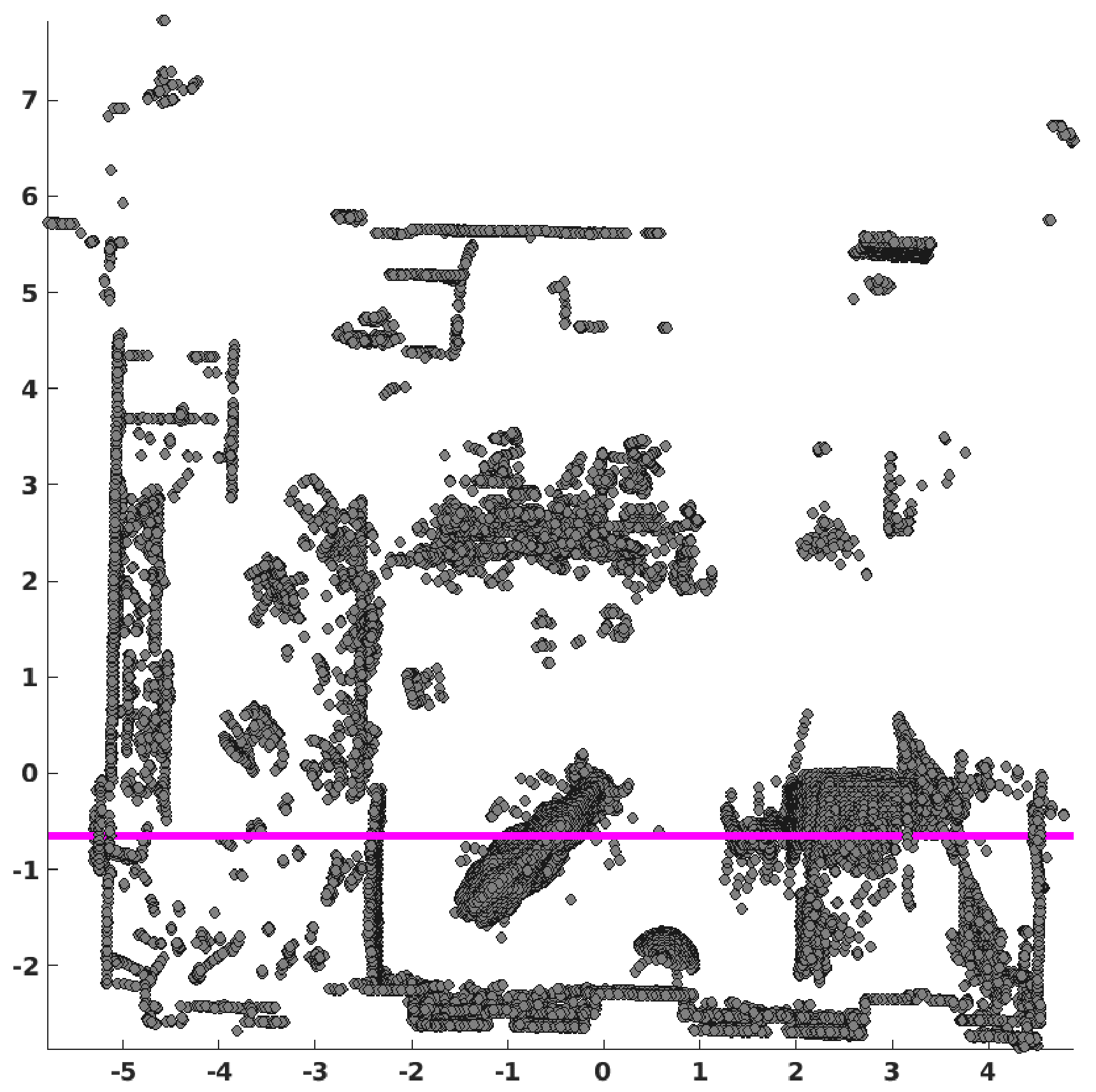} \label{ptcloud}}
        \subfigure[Obstacle GPDF ]{
        \includegraphics[width=30mm]{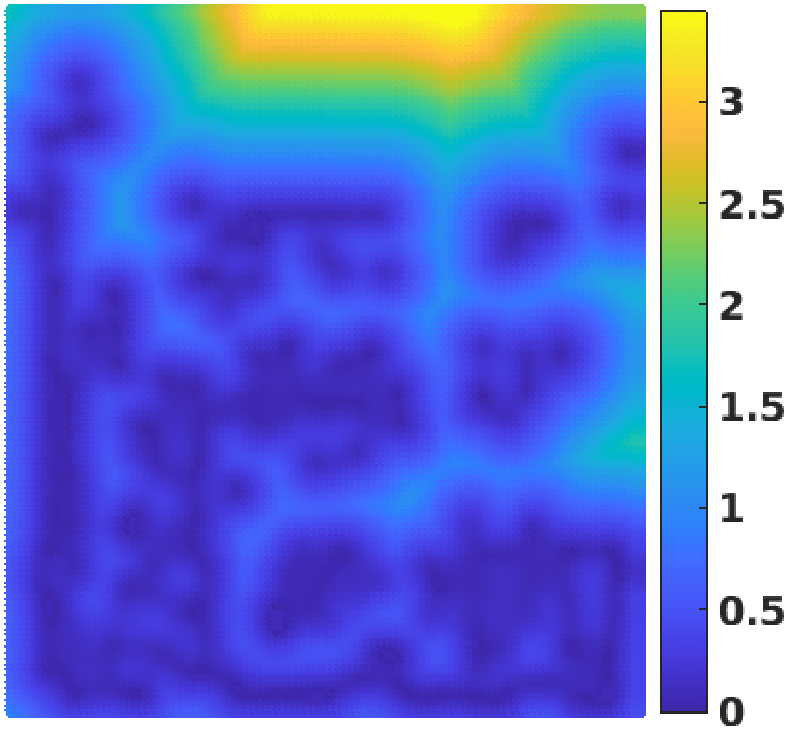}\label{obstacle}}
        \subfigure[Ground GPDF]{
        \includegraphics[width=35mm]{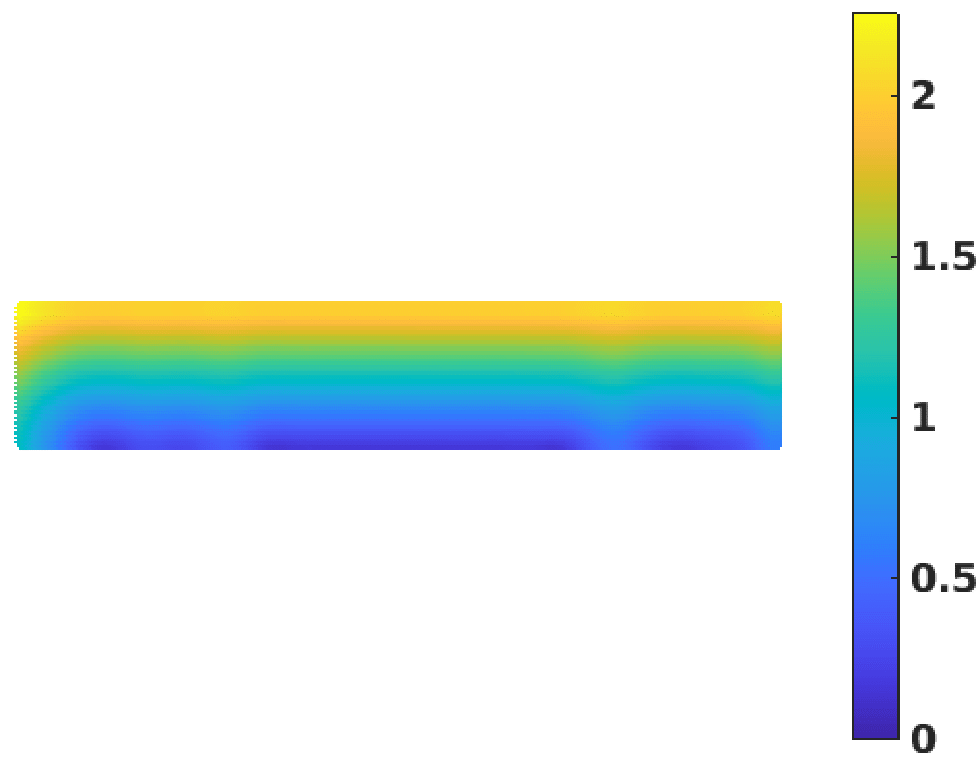} \label{ground}}
    \caption{(a) Point cloud map of the \textit{Cow and lady} dataset, the ground points and any points above $2$ m in height have been removed. 
(b) Horizontal slice (top-view) of the Obstacle GPDF showing the distance to the nearest obstacle points from a horizontal slice taken $1$ m above the ground. (c) Vertical slice of the Ground GPDF along the magenta line shown in (a).
}
    \label{gpdf_two}
    \vspace{-5ex}
\end{figure}

\subsubsection{Ground GPDF.}
Given the segmented ground points set $\mathbf{X}_g = \{ \mathbf{x}_{g_1}, \mathbf{x}_{g_2}, \ldots, \mathbf{x}_{g_n} \}$ $ \in \mathbb{R}^3$, let us first formulate the Ground GPDF. As introduced in Section \ref{background}, we first obtain the occupancy, ${o_g(\mathbf{x}_g)\sim \mathcal{GP}(0,k_{o_g}(\mathbf{x},\mathbf{x'}))}$. By applying the reverting function in \eqref{reverting_func}, we then obtain the Ground GPDF, and the corresponding distance function $d_g(\mathbf{x})$. \Rev{We choose the Squared Exponential (SE) kernel because it is smooth and widely utilised in the literature, but more importantly because it has a closed-form solution for the reverting function}, 
\begin{equation}
d_g=r(o_g)=\sqrt{-2l^2 \log(\frac{o_g}{\sigma^2})}.
\label{reverting_func}
\end{equation}

\Rev{Note that the reverting function provides an accurate approximation of the true distance field and the same kernel formulation is applicable regardless of the map. Furthermore, since it is analytically differentiable, it can generate reliable gradient values required for trajectory optimisation. In our experiments, we set the lengthscale, \textit{l}, to be approximately 2 times the average distance between neighbouring points on the same surface. }

In Fig. \ref{ground}, we show an example of the Ground GPDF for the \emph{Cow and lady} dataset~\cite{Voxblox}. 
A vertical slice of this Ground GPDF is taken at $-0.65m$ along the Y-axis for visualisation. 
    \vspace{-2ex}

\subsubsection{Obstacle GPDF.}
After obtaining the Ground GPDF, we aim to find the obstacle set $\mathbf{X}_o\in \mathbf{X}_n$, that considers all obstacles that are within the height of the mobile system, e.g., overhanging objects and obstacles above the ground. With the Ground GPDF, let us query the distance to the ground of all the non-ground points and extract those that are within the height limit of the mobile system $h_r$.
These points become the obstacle set, $\mathbf{X}_o = [\mathbf{x}_{o_1}, \mathbf{x}_{o_2}, \ldots, \mathbf{x}_{o_n}] \in \mathbb{R}^2$.
We then project these points to a 2D plane
and build the Obstacle GPDF. 

We use the same \Rev{kernel} function as for the Ground GPDF 
in \eqref{reverting_func}, but now with ${\mathbf{X}_o}$ to obtain $k_{o_o}(\mathbf{x}_o,\mathbf{x}'_o)$ and its reverting function $r(o_o)$ to estimate the distance function $d_o (\mathbf{x}_o)$ to the obstacles for a given mobile system. In Fig. \ref{obstacle}, a horizontal slice of the Obstacle GPDF, taken at $1$m is shown as an example. 
Note that this Obstacle GPDF calculates only the 2D Euclidean distance in the X-Y plane since we have projected the 3D points into 2D to get the obstacle set in simplified form. 

\begin{figure}[h]
    \centering
    \subfigure[3D view]{
        \includegraphics[width=50mm]{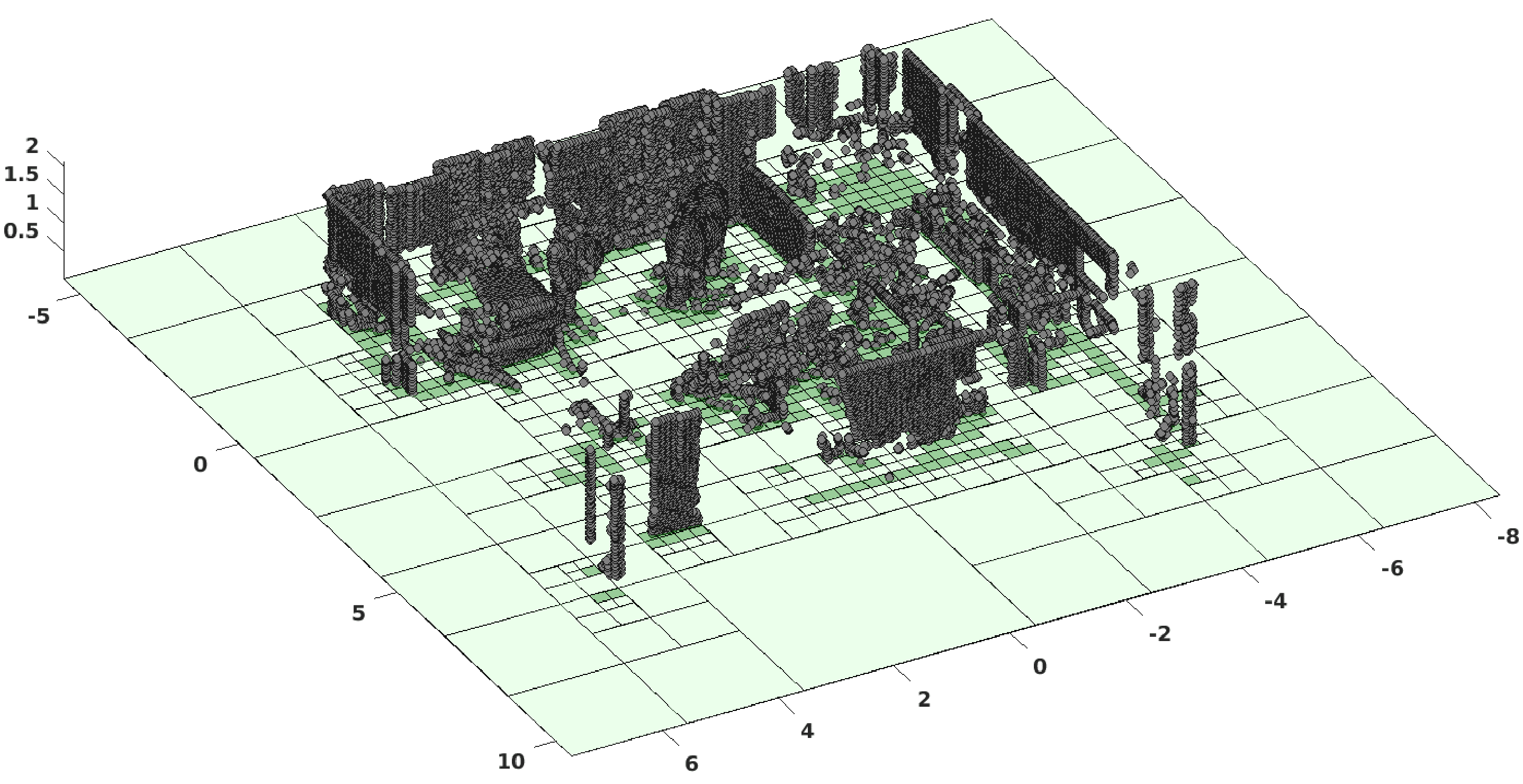}}
        \subfigure[Top view]{
        \includegraphics[width=45mm]{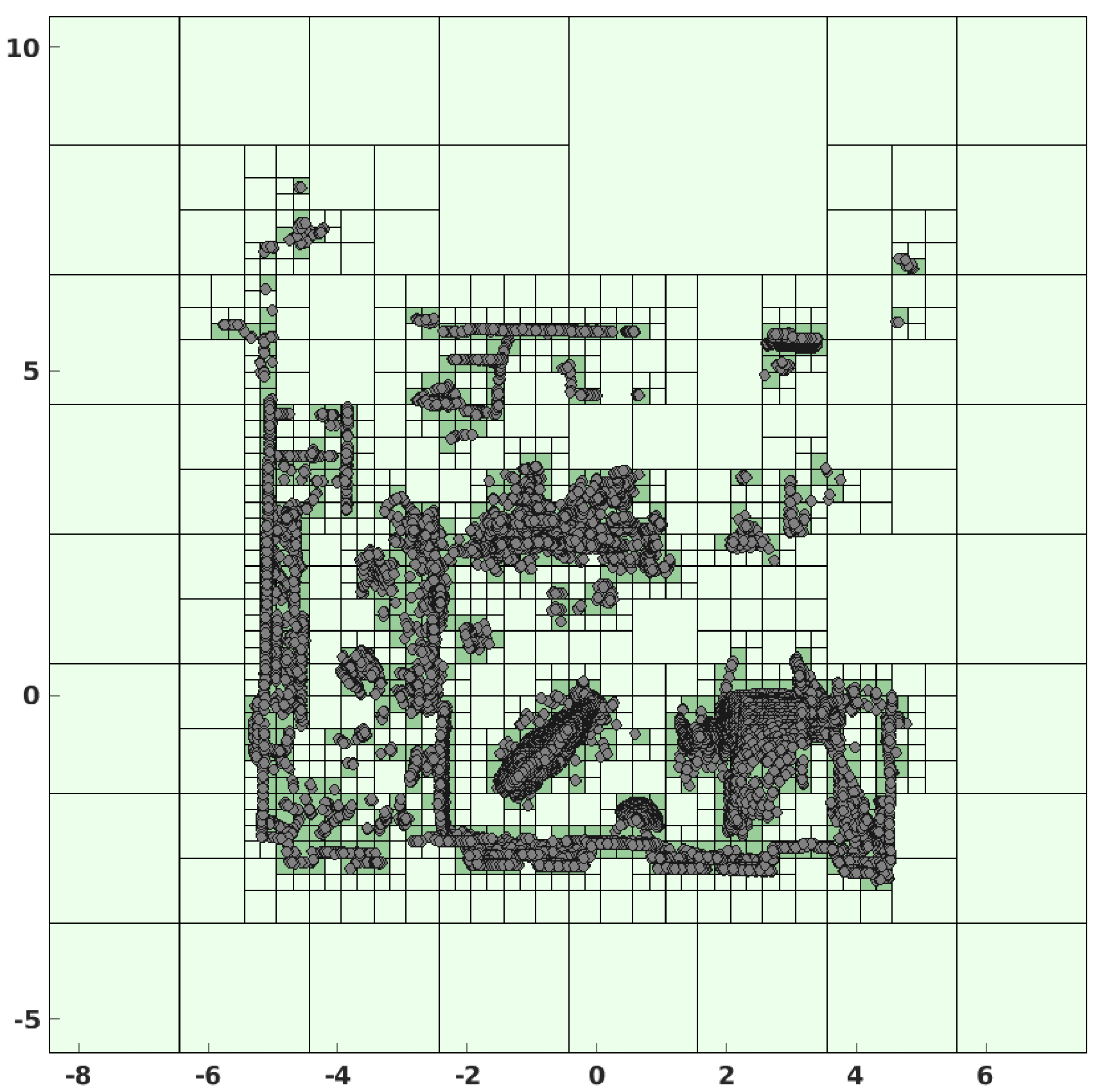}}
    \caption{Quadtree formulation on the \textit{Cow and lady} dataset. Dark green boxes indicate cells containing points (occupied), and light green boxes denote empty cells (free space). In (a), the quadtree is shown with obstacle points plotted in 3D. Followed by the quadtree with projected 2D obstacle points in (b)}
    \label{quadtree_formulate}
    \vspace{-5ex}
\end{figure}

\subsubsection{Quadtree Formulation.}
Given the obstacle points, $\mathbf{X}_o \in \mathbb{R}^3$, we only consider the points in X and Y coordinates, thus making the 3D point cloud, a 2D one projected on X-Y plane. We use this squashed point cloud $\mathbf{X}_o(\mathbf{x},\mathbf{y}) \in \mathbb{R}^2$, to formulate a quadtree structure. 
To ensure the preservation of free space information, we keep track of the empty cells in the tree and their connectivity information. 
In Fig. \ref{quadtree_formulate}, we show an example of the quadtree formulation with the \textit{Cow and lady} dataset. The dark green cells indicate the occupied cells with points from the point cloud. The light green boxes denote the empty cells and, thus, the free space with no obstacles. By monitoring vertex and edge connections of the empty nodes, we create a 2D connectivity graph of the free space. This sparse graph, representing only the free space information, is given to an $A^*$ search which generates a valid 2D initial trajectory for further optimisation. This formulation allows us to reason our planning in free space, constructed in a multiresolution format.

 \subsection{Trajectory Optimisation in 3D}
 We choose CHOMP as the trajectory optimisation method to be used. The trajectory \( \mathbf{x}^r \) consists of \( Q \) waypoints $\{\mathbf{x}_1^r, \mathbf{x}_2^r, \ldots, \mathbf{x}_Q^r\} \in \mathbb{R}^D $. Thus, the complete trajectory is in \(\mathbb{R}^{Q \times D}\). We follow the original formulation in \eqref{chomp_obj} to enable collision-free, smooth trajectories in 3D space and add an extra term to generate feasible paths that remain at a constant distance to the ground. 
 


The additional cost,  
which enforces the ground constraint is given by,
\begin{equation}
c_g(\mathbf{x}_i^r) = \frac{1}{2*h_r} \left( d_g(\mathbf{x}_i^r) - h_r \right)^2\,,
 \end{equation}
where $h_r$ is the value of the maximum height of the ground-based system and $d_g(\mathbf{x}_i^r)$ is the distance to the ground from the $i$-th waypoint $\mathbf{x}_i^r$ in the trajectory.
The final objective function for the optimisation is given by,
 \begin{equation}
C[\mathbf{x}^r] \equiv  \sum_{i=1}^{Q}  ( \Rev{ \lambda_s}\frac{1}{2} \|\mathbf{\dot{x}}_i^r\|^2 +  \lambda_o c_o(\mathbf{x}_i^r) \|\mathbf{\ddot{x}}_i^r\| +  \lambda_g c_g(\mathbf{x}_i^r))
\end{equation}
with \Rev{ $\lambda_s$}, $\lambda_o$ and $\lambda_g$ as the weights for \Rev{smoothing cost}, obstacle avoidance and \Rev{ground} constraint. \Rev{ These weights need to be tuned according to the map's specific features. However, in all cases, \Rev{$\lambda_s$} should be set significantly lower than the other two weights to prioritize obstacle avoidance and maintaining the ground constraint. This adjustment is crucial for ensuring the feasibility of the optimized trajectory. Additionally, the magnitudes of $\lambda_o$ and $\lambda_g$ should be relatively close to one another to maintain balance, ensuring that both constraints are adequately satisfied.} 

The term $c_o(\mathbf{x}_i^r)$ is the obstacle avoidance cost detailed in \eqref{obs_cost}. In our pipeline, we use the Obstacle GPDF inference $d_o$ for each waypoint in $\mathbf{x}^r$. As we are working with an unsigned distance field, all distance values remain positive. Thus, the condition for negative distances in \eqref{obs_cost} is not taken into account.

For the $c_g(\mathbf{x}_i^r)$ cost, the distance to the ground is $d_g(\mathbf{x})$ using the Ground GPDF.  
This not only ensures the trajectory is at a constant height but also ensures handling any uneven traversable terrain, such as a podium or a step. 
\vspace{-3ex}

\subsubsection{Initial Trajectory.}
By using our obstacle projection quadtree representation and the free cell connectivity graph described in Section~\ref{sec:rep}, we can obtain a good initial guess with an $A*$ search that guarantees convergence in the optimisation. Note that we use the Euclidean distance between two nodes, as the heuristic cost to be used in the $A*$ search.\\






\vspace{-3ex}

\section{Evaluation}
\MU{In this section, we evaluate the proposed approach and showcase our results. First, we discuss the datasets used in these experiments. We then validate the importance of the two GPDFs in our proposed framework. Afterwards, we benchmark the approach with three different methods to evaluate its performance. We discuss the benchmarks, the experiments performed and their respective comparative results. Furthermore, we validate the performance across multiple ground-based systems. Note that to enhance readability, most visualisations do not show the ground floor, except for elevated steps. Finally, we provide a quantitative evaluation on safety and feasibility for the planned trajectory. }
     
     
\subsection{Implementation Details}
\subsubsection{Datasets.}
We use \Rev{three} publicly accessible real-world datasets, \textit{Cow and lady} \cite{Voxblox}, \textit{Stanford 2D-3D} \cite{stanford2017} \Rev{and \textit{DARPA Subterranean challenge} \cite{tranzatto2022cerberus}} datasets. For the \Rev{\textit{Cow and Lady dataset}}, a PLY point cloud, which was produced by combining several scans with a Leica MS50 professional laser scanner is used as the input.\\
The \textit{Stanford 2D-3D} dataset comprises six indoor areas. We used the raw point cloud data from two office environments: one from Area 1 (Office 1) and one from Area 4 (Office 10). To demonstrate our method's ability to handle overhanging obstacles, we added a round, ball-shaped object positioned 0.7 meters above the ground to the \textit{Office 1} dataset. This object represents anything hanging from the ceiling. Additionally, we included a step sized of $1$m $\times 1.2$m $ \times 0.35$m on the side of the room. We consider this as a terrain traversable by humans and quadruped robots. 
For the \textit{Office 10} dataset, we used directly the raw point cloud data.

\Rev{The \textit{DARPA Subterranean challenge} dataset contains maps of complex environments from both urban and underground areas. For our work, we used point cloud data of two distinct maps: the \textit{urban circuit} and the \textit{cave circuit}. The first one includes detailed urban features like staircases, while the other has rough, uneven terrains typical of subterranean landscapes.}

\subsubsection{Benchmarks.}
We propose to benchmark against vanilla CHOMP with a single 3D GPDF, the A* based on the 2D graph from Quadtree and a PRM algorithm. Since the A* trajectory is projected in 2D, we offset it to the X-Y plane that crosses the start location at the maximum height of the system. We opt for a similar approach for PRM as well, where from a 2D projection of the map, we find the binary occupancy and, subsequently, a path from start to goal before offsetting it at the maximum height. \Rev{Note that, in the case of the \textit{urban circuit} dataset, there are no obstacle points. Thus, both A* and PRM output a straight line from start to goal.} To replicate sensor noises, we add a Gaussian noise with $\sigma = 5$cm to the point cloud data for all the algorithms in consideration. Furthermore, for evaluating our framework across different robotic systems, we take into consideration the exact dimensions of the system and consider a cylindrical model to evaluate safety and ground constraints.
\vspace{-3ex}
\begin{figure}
    \centering
         \setlength{\belowcaptionskip}{0pt}  

         \subfigure[3D view]{
        \includegraphics[width=45mm]{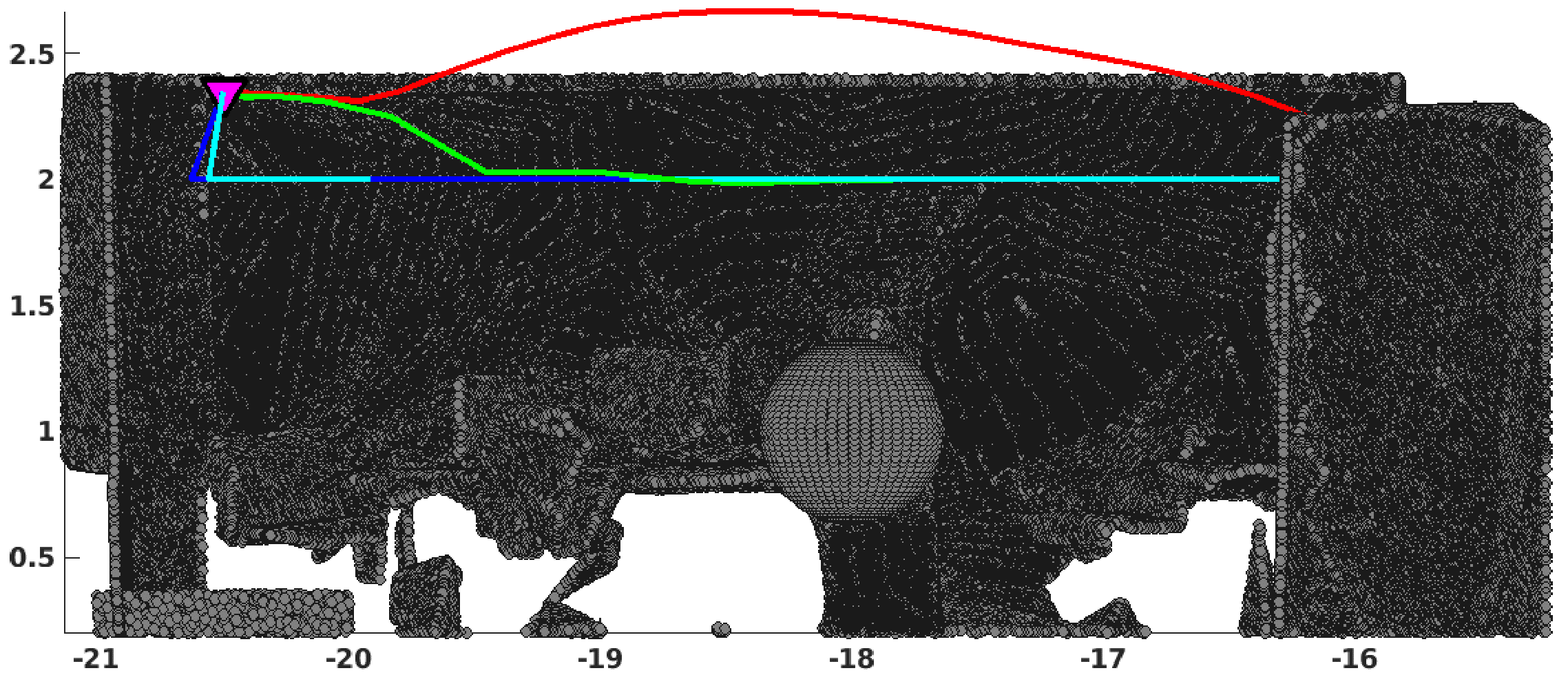 }}
        \subfigure[Top view]{
        \includegraphics[width=70mm]{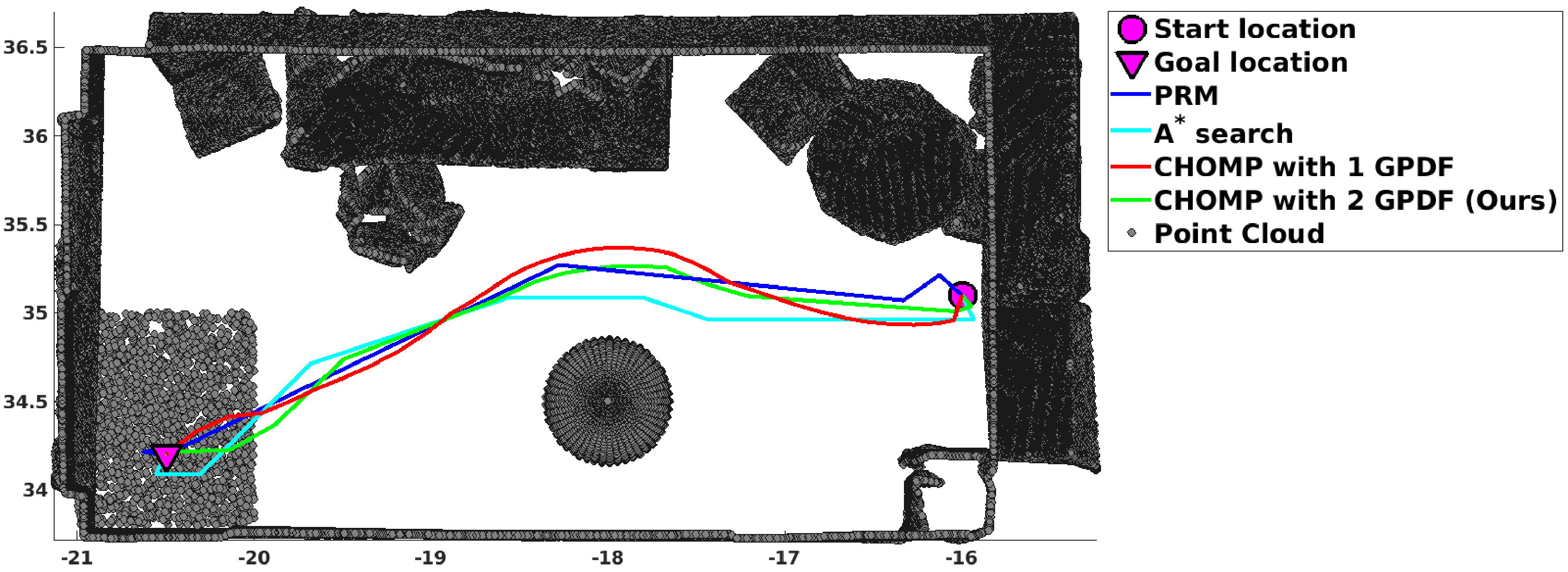}}
 
    \caption{Trajectory generated with PRM (blue), $A^*$ (cyan), CHOMP using a single trained GP for distance related query (red line) and CHOMP using two GPDFs (green line),  on the \textit{Stanford 2D-3D's Office 1} dataset. For better visualisation, the wall on the side of the ball has been removed in the 3D view (a).}
    \label{stanford_ball_tool}
    \vspace{-8ex}
\end{figure}

\subsection{Validating Our Representation}
\begin{figure}
    \centering
     
         \subfigure[3D view (\textit{urban circuit} dataset)]{
        \includegraphics[width=55mm]{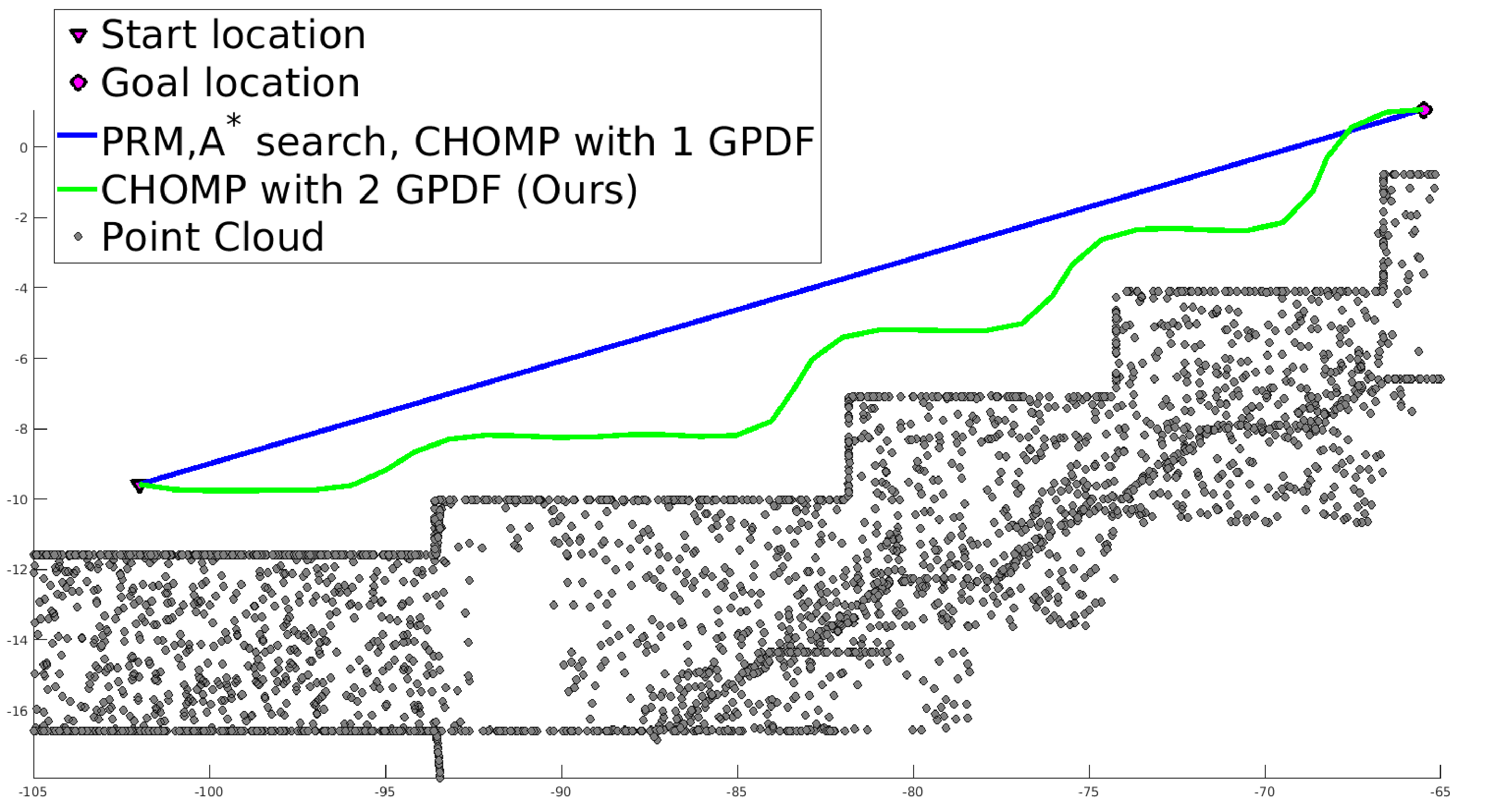}}
        \subfigure[3D view (\textit{cave circuit} dataset)]{
        \includegraphics[width=60mm]{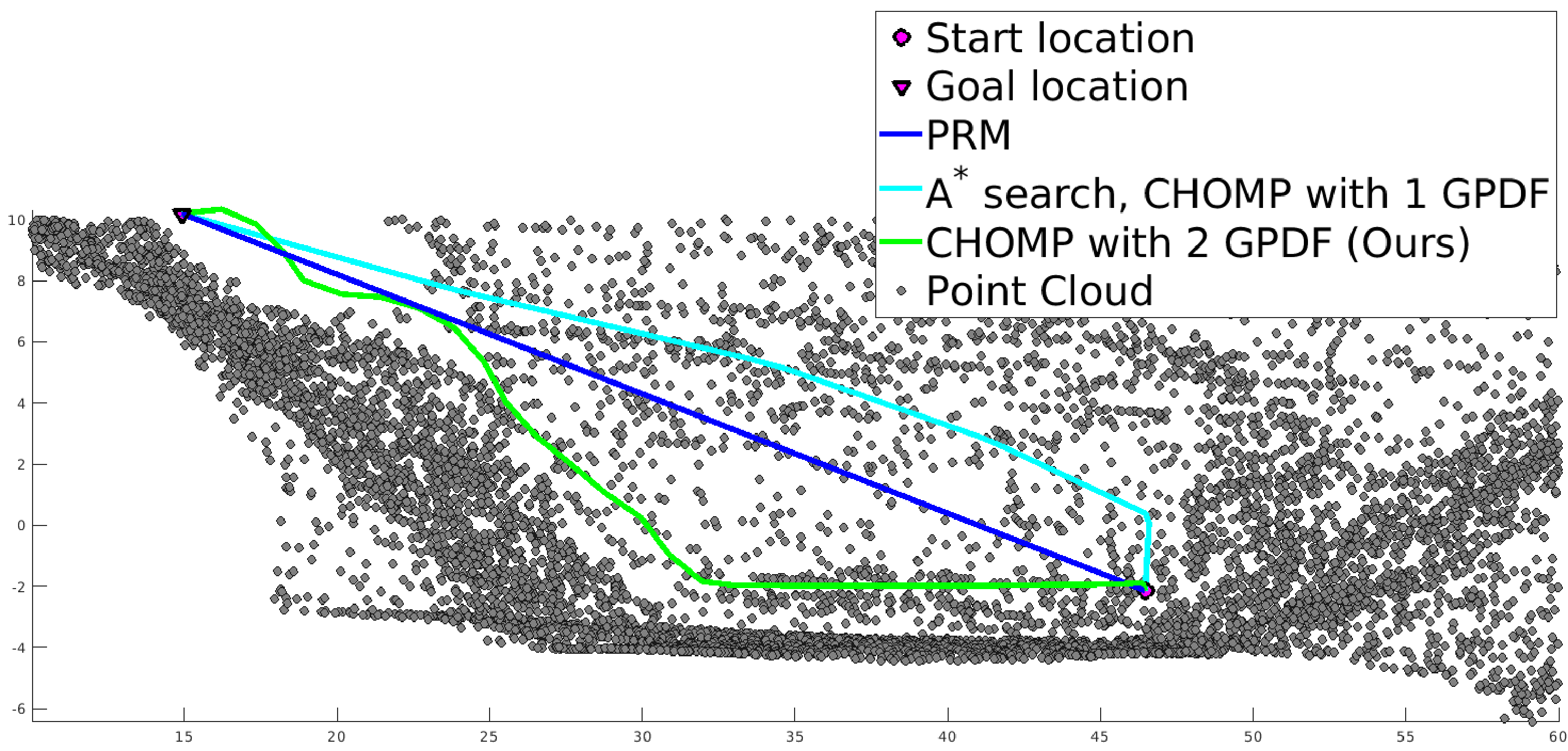}}
 
    \caption{\Rev{(a) Trajectory generated with PRM, $A^*$ and CHOMP using single GPDF (blue) and CHOMP using two GPDFs (green),  on the \textit{urban circuit} dataset (b)Trajectory generated with PRM(blue), $A^*$ and CHOMP using single GPDF (cyan) and CHOMP using two GPDFs (green),  on the \textit{cave circuit} dataset }}
    \label{darpa}
    \vspace{-5ex}
\end{figure}

Since our scene representation can cope with uneven terrain and elevated surfaces, we compare the results on a trajectory with the end goal on top of an elevated platform. We use our modified \emph{Office 1} dataset to demonstrate the soundness of our framework. Fig.~\ref{stanford_ball_tool} clearly shows that compared to others, ours is the only one that leads to a feasible trajectory while maintaining the ground-constrained that correctly accounts for the step. Both A* and PRM fail to handle the elevation. CHOMP with one GPDF fails to maintain the desired height.


\Rev{To evaluate our approach in a more challenging terrain, we use the DARPA Subterranean challenge datasets. From Fig. \ref{darpa}, it is evident that ours is the only one capable of maintaining the ground constraints. Our approach is robust enough to successfully produce feasible trajectories, varying from long staircases to subterranean maps with irregular elevations. In contrast, the other methods not only fail to account for uneven terrain in their trajectories but also exceed the system's height limits, causing them to lose consistent contact with the ground.} 

\vspace{-3ex}

\subsection{Comparative Results on Datasets}
\vspace{-3ex}

\begin{figure}
    \centering
     
       \subfigure[3D view]{
        \includegraphics[width=40mm]{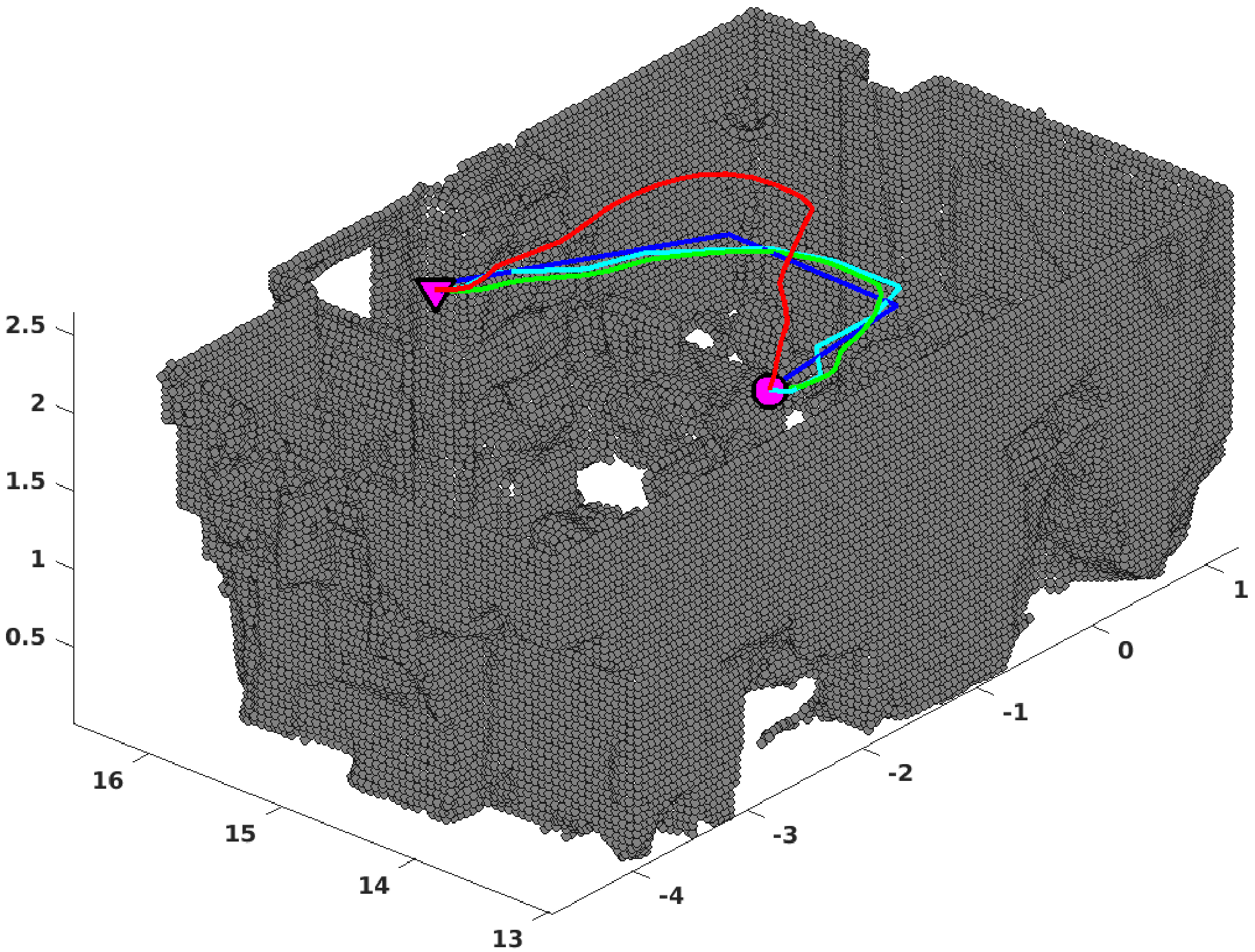 }}
        \subfigure[Top view]{
        \includegraphics[width=65mm]{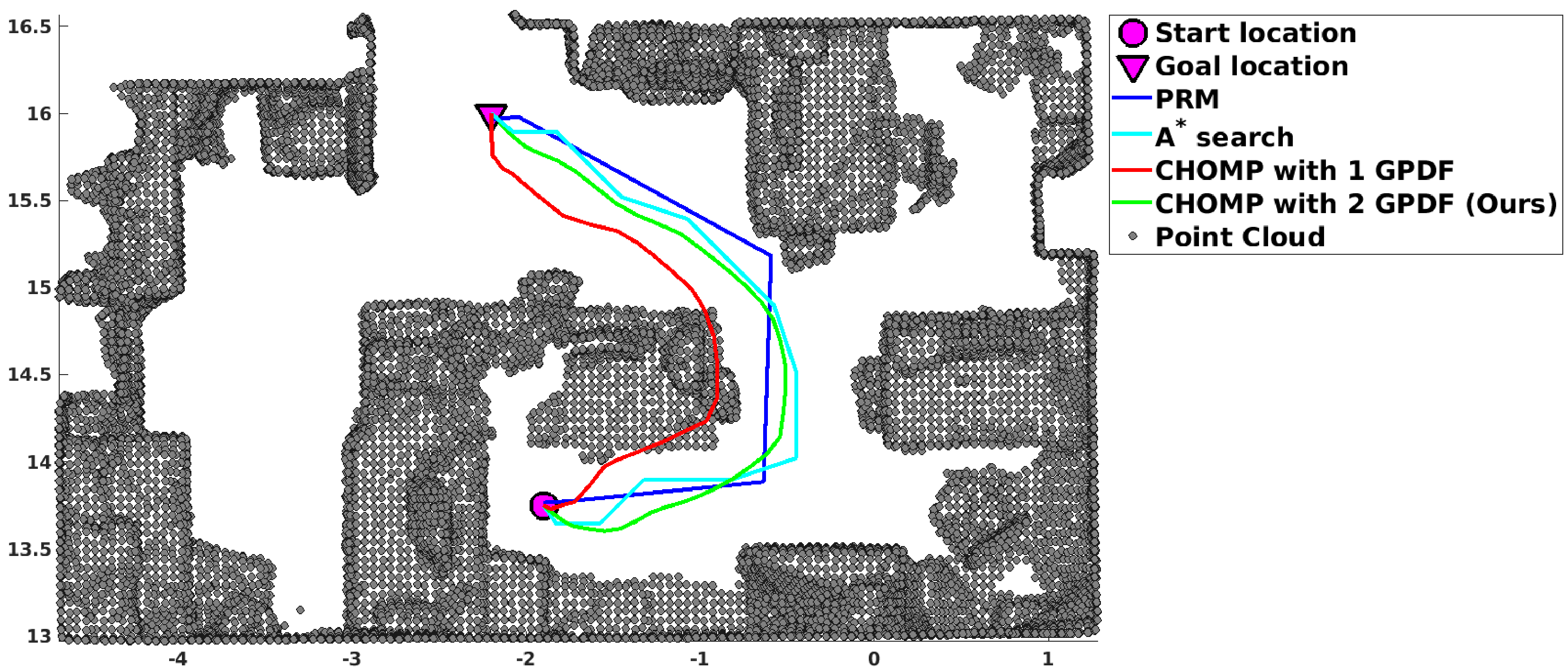}}
     
    \caption{Trajectory generated with PRM (blue), $A^*$ (cyan), CHOMP using a single trained GP for distance related query (red line) and CHOMP using 2 GPDFs (green line), on the \textit{ Stanford 2S-3D's Office 10} dataset.}
    \label{stanford}
    \vspace{-5ex}
\end{figure}
Fig. \ref{stanford} shows how our approach fairs compared to the other approaches in the cluttered \textit{Office 10} dataset. Even though all the approaches provide collision-free trajectories, both PRM and A* fail to consider any notion of safety and would very likely collide at a few critical points, given the dimensions of the agent in consideration. The traditional CHOMP, augmented with a single 3D GPDF, goes over the obstacles and fails to satisfy the ground constraints. In contrast, ours takes the safest route, maintaining a desired distance as well as height to ensure the path is safely traversable for the system.

\begin{figure}
    \centering
    
         \subfigure[3D view]{
        \includegraphics[width=55mm]{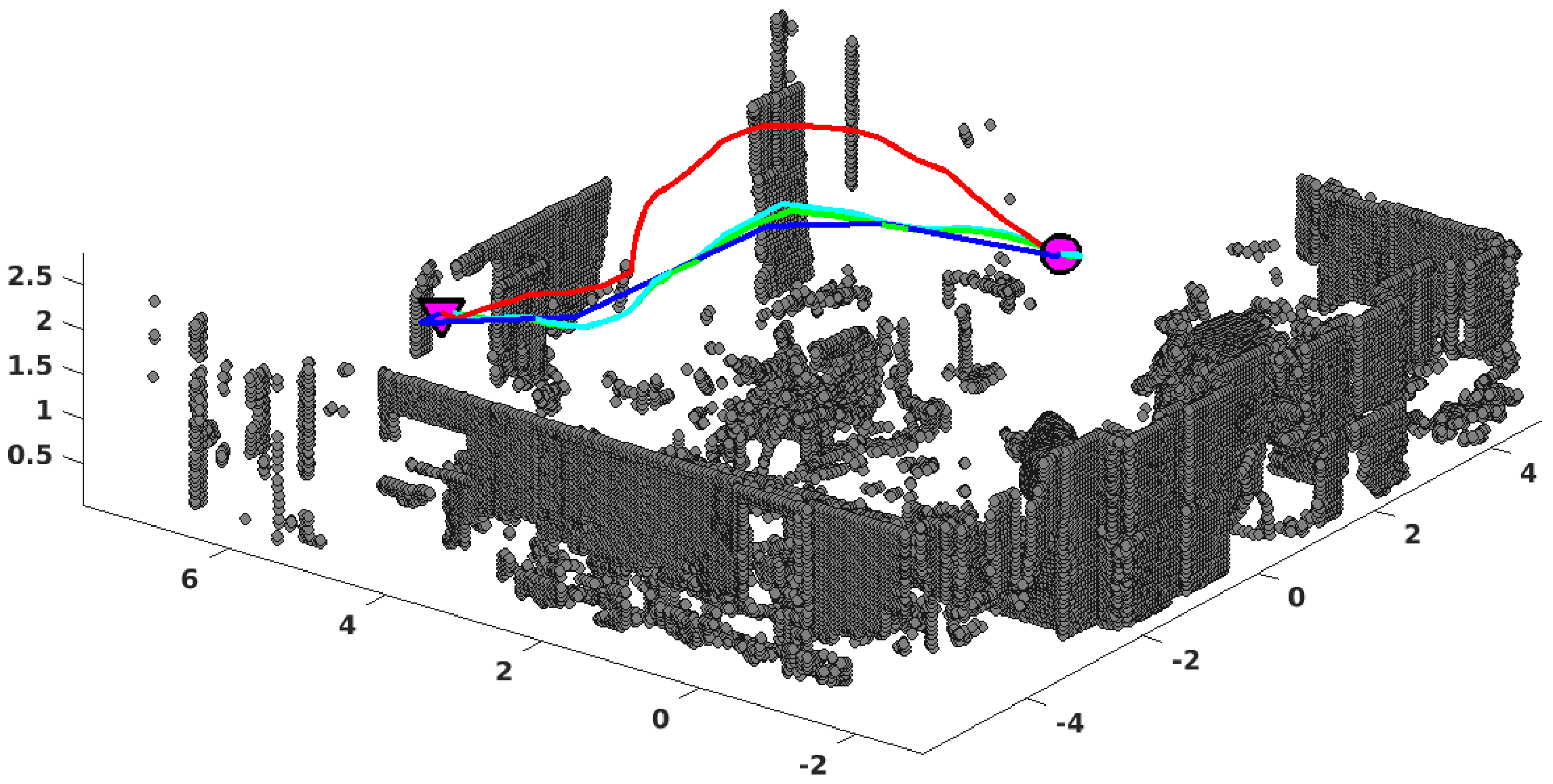}}
        \subfigure[Top view]{
        \includegraphics[width=60mm]{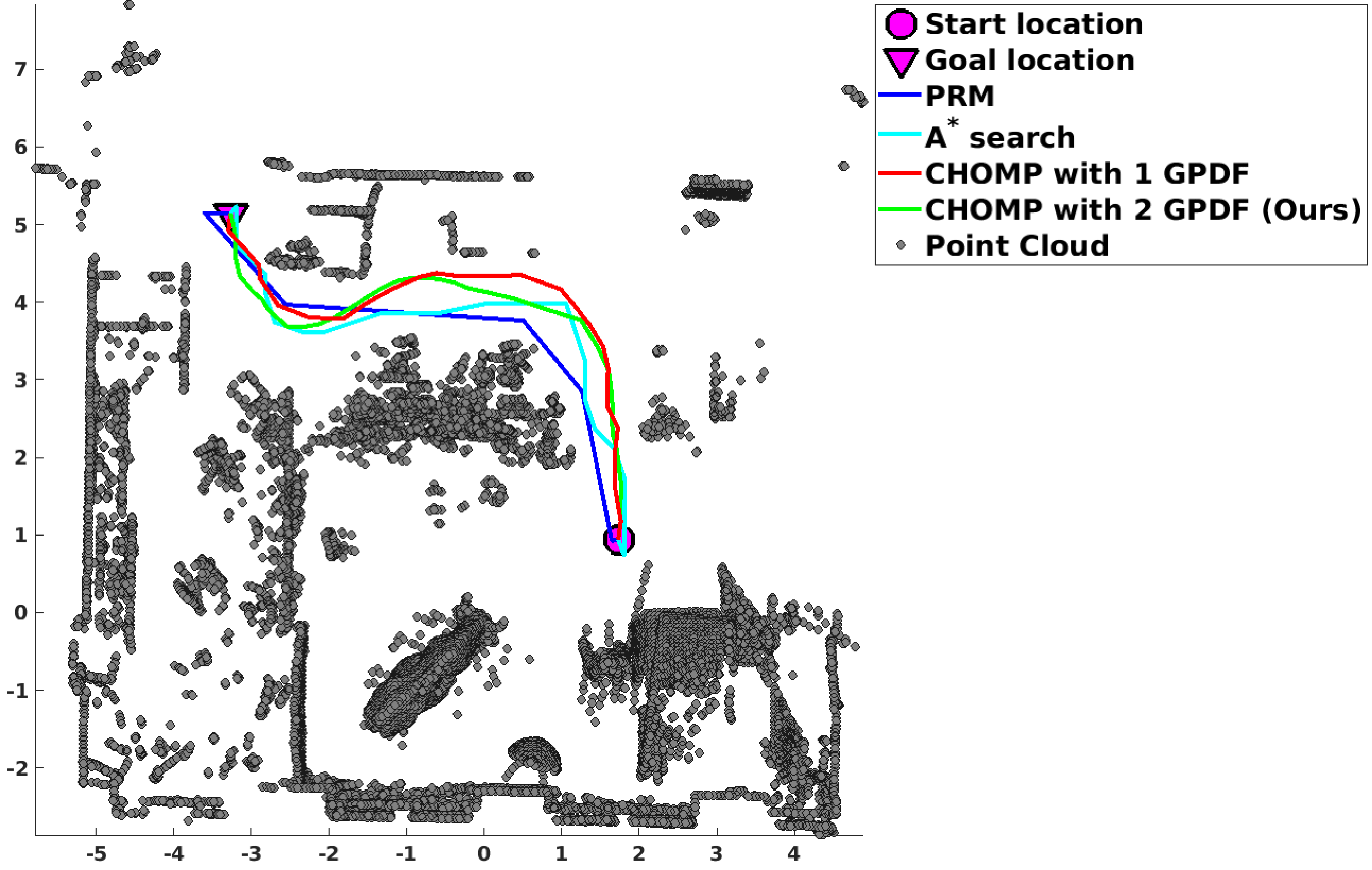}}
    \caption{Trajectory generated with PRM (blue), $A^*$ (cyan), CHOMP using a single trained GP for distance related query (red line) and CHOMP using 2 GPDFs (green line),   on the \textit{Cow and lady} dataset.}
    \label{cow_and_lady}
    \vspace{-10ex}
\end{figure}

The results are similar in the \textit{Cow and lady } dataset as well (Shown in Fig. \ref{cow_and_lady}). Both PRM and A* contain critical points that would lead to a collision with a real robot, failing to maintain a desired distance. Unlike the previous, the single GPDF with CHOMP is able to find a collision-free path, but it does not maintain any consistent height; rather, it gives an infeasible optimised trajectory that does not keep contact with the ground. Compared to these, ours is much safer, maintaining a safe distance and constricting the trajectory at the desired height from the ground.


\vspace{-3ex}

\subsubsection{Trajectory Analysis for Different Mobile Systems.}
\begin{figure}
    \centering
    \setlength{\belowcaptionskip}{0pt}  
    \subfigure[3D view]{
        \includegraphics[width=45mm]{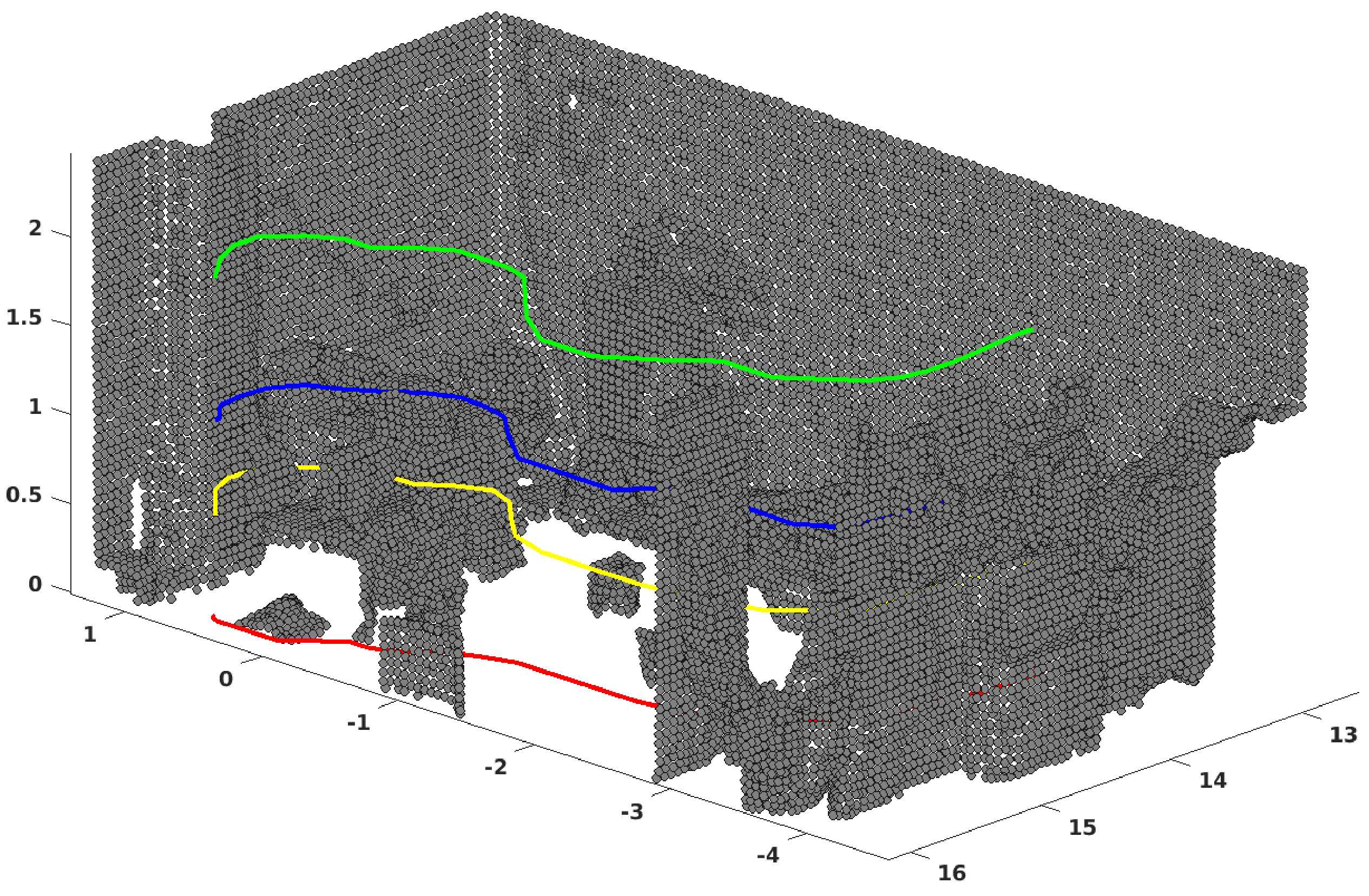 }}
        \subfigure[Top view]{
        \includegraphics[width=52mm]{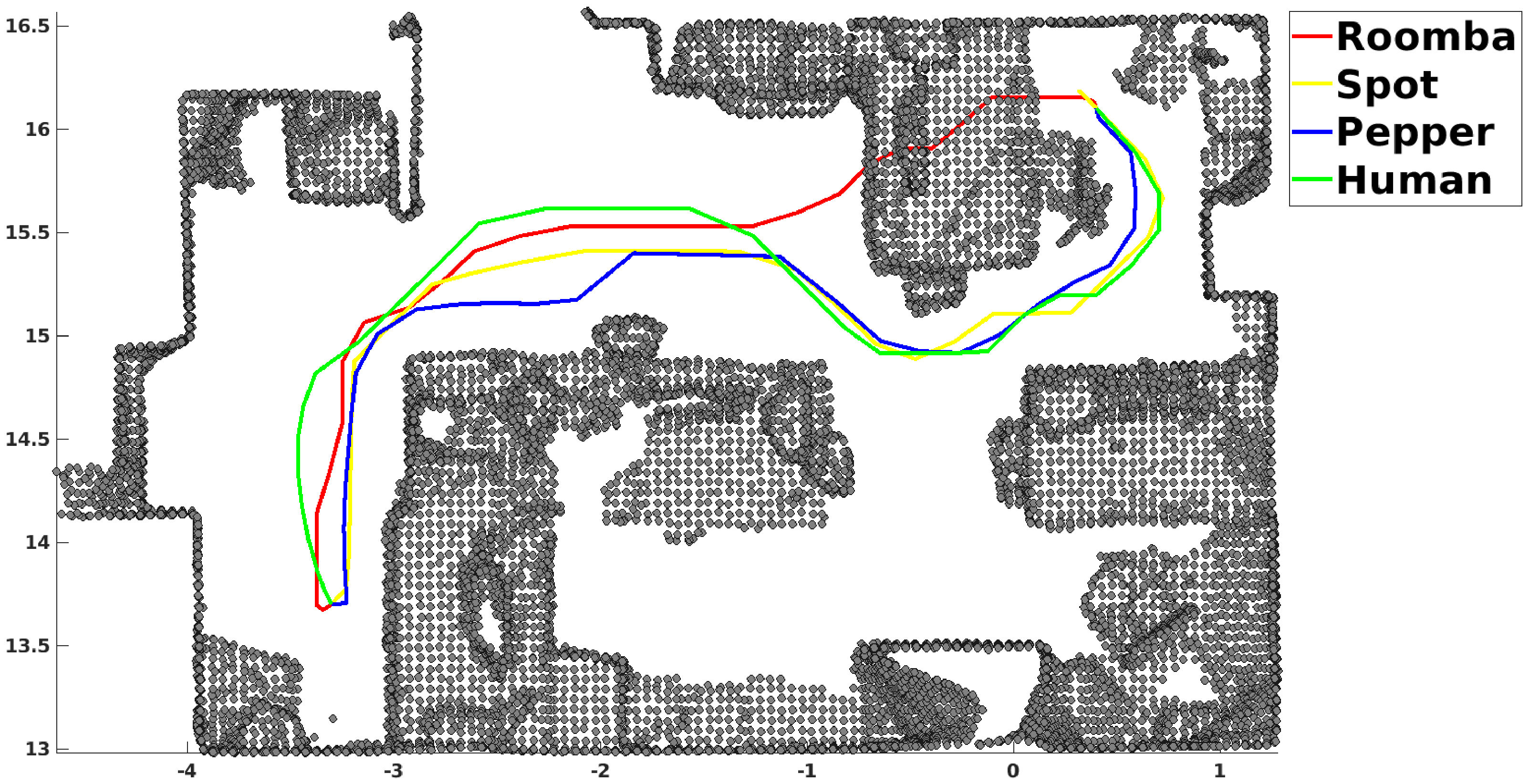}}
    \caption{ Comparative trajectories generated with our method for different agents: Roomba (red), Boston Dynamics Spot (yellow), Pepper bot (blue), and human (green), with maximum heights approximately at 0.1 m, 0.7 m, 1.2 m, and 2 m, respectively. 3D view (a) and top view (b) are shown, on the \textit{ Stanford 2S-3D's Office 10} dataset.}
    \label{all_trajectory}
\end{figure}

To showcase the potential of our approach to constrict the height of the trajectory at the maximum height, as well as maintain a safe distance depending on the agent's dimensions (Table \ref{table:dims}), we demonstrate a long trajectory in the \textit{Office 10} dataset (Fig. \ref{all_trajectory}), and how it behaves in the case of agents with different dimensions. 
\vspace{-3ex}

\begin{table}
\centering
\begin{tabular}{|c|c|c|c|}
\hline
Ground-based system & Height ($m$) & Width ($m$) & Length ($m$)\\
\hline
Roomba &   0.1& 0.34 & 0.35 \\
\hline
Spot &  0.7 &  0.19 & 1.1\\
\hline
 Pepper  &  1.2&  0.48& 0.42   \\  
\hline
Human & 2 & 0.5 &  0.32 \\
\hline
\end{tabular}
\caption{Dimensions of different ground-based systems}
\label{table:dims}
\vspace{-8ex}
\end{table}

In all cases, our approach is able to find a solution that maintains the required distance from obstacles. An interesting observation is of the Roomba's. Its dimensions allow it to traverse under the couch in contrast to the longer route the other agents took, meaning the trajectory is optimal for the agent as well.

\subsubsection{Safety and Feasibility Evaluation.}
  
In the absence of an established metric for safety, we propose to run a simulation 100 times with randomly initialised start and goal locations over the whole map for each algorithm. We propose two separate experiments, one to measure success rates of giving collision-free trajectory, and another to measure the feasibility of the trajectory. We choose to conduct the experiment with two systems with vastly different dimensions, Human and Roomba. For the success rates, we ensure that each waypoint is free from collision given the cylindrical model of the respective system. 

\begin{table}
\setlength{\belowcaptionskip}{0pt}
\centering
\begin{tabular}{|c|c|c|c|c|c|}
\hline
\multirow{2}{*}{Ground-based system} &\multirow{2}{*}{Dataset} & \multicolumn{4}{c|}{Success rate of different methods (\%)} \\
\cline{3-6}
 & & PRM & A* & CHOMP with one GPDF & \textbf{Ours}\\
\hline

\multirow{3}{*}{Human}&\textit{Cow and lady}& 54 & 62 & \textbf{96} & \textbf{96}\\
\cline{2-6}
&\textit{Office 10}& 28& 54 & 93 & \textbf{96} \\
\cline{2-6}
&\textit{Office 1}& 60& 71 & 98 & \textbf{100} \\
\cline{2-6}
&\Rev{\textit{Cave circuit}}& \Rev{87}& \Rev{96} & \Rev{99} & \Rev{\textbf{100}} \\
\hline

\multirow{3}{*}{Roomba}&\textit{Cow and lady}& 72 & 76 & \textbf{100} & \textbf{100}\\
\cline{2-6}
&\textit{Office 10}& 58& 68 & 98 & \textbf{100} \\
\cline{2-6}
&\textit{Office 1}& 81& 76 & \textbf{100} & \textbf{100} \\
\hline
\end{tabular}

\caption{Success rates across different datasets for collision-free trajectory}
\label{safe_rate}
\vspace{-5ex}
\end{table}

In the case of the feasibility study, we check if every waypoint is within the desired height from the ground allowing an error margin of $\pm10\%$, e.g., for humans, the height limit is $1.8-2.2 m$ from the ground. It should be noted that since the Roomba cannot climb the platform, the feasibility study for Roomba on the \emph{Office 1} dataset we do not include the platform in the map. \Rev{Also, for \textit{cave circuit} dataset, we have done all the quantitative analysis for humans only.}
\vspace{-3ex}
\begin{table}
\setlength{\belowcaptionskip}{0pt}  

\centering
\begin{tabular}{|c|c|c|c|c|c|}
\hline
\multirow{2}{*}{Ground-based system} &\multirow{2}{*}{Dataset} & \multicolumn{4}{c|}{Success rate of different methods (\%)} \\
\cline{3-6}
 & & PRM & A* & CHOMP with one GPDF &  \textbf{Ours} \\
\hline
\multirow{3}{*}{Human}&\textit{Cow and lady}& \textbf{100} & \textbf{100}& 47 & \textbf{100}\\
\cline{2-6}

&\textit{Office 10}& \textbf{100}& \textbf{100}& 19 & \textbf{100} \\
\cline{2-6}
&\textit{Office 1}& 83& 78 & 28 & \textbf{99} \\
\cline{2-6}
&\Rev{\textit{Cave circuit}}& \Rev{13}& \Rev{8} & \Rev{7} & \Rev{\textbf{97}} \\
\hline
\multirow{3}{*}{Roomba}&\textit{Cow and lady}& \textbf{100} & \textbf{100} & 23 & \textbf{100}\\
\cline{2-6}
&\textit{Office 10}& \textbf{100}& \textbf{100} & 20 & \textbf{100} \\
\cline{2-6}

&\textit{Office 1}& 86& 83 & 6 & \textbf{100} \\
\hline
\end{tabular}

\caption{Success rates across different datasets for feasible trajectory}
\vspace{-5ex}
\label{tab:feas_rate}
\end{table}

The results of the success rates of generating collision-free trajectory are shown in Table \ref{safe_rate}. Across all the datasets, ours performs significantly better than both A* and PRM. 
It should be noted that CHOMP with a single GPDF performs similarly to ours, as it manages to avoid collisions consistently, often going above the obstacles. This is exemplified in the feasibility study (Table \ref{tab:feas_rate}), which showcases that the same is often unable to maintain the waypoints within the ground constraints of the system. Especially in the case of the Roomba, where there is not much room for error in terms of feasibility, the single GPDF struggles to produce valid trajectories. In contrast, ours produces valid and feasible trajectories and shows the best performance among all the methods throughout the datasets. Also, it is also apparent that with smaller dimensions, our approach reaches a 100\% success rate in generating both safe and feasible trajectories.\\
\Rev{\quad To evaluate the system's behaviour on uneven terrain, we calculate the average distances from both the ground and obstacles. Experiments are conducted considering human subjects using the \textit{office 1} and \textit{cave circuit} datasets, as these feature uneven/rough terrains with obstacles. Table \ref{tab:avg_distances} shows that among all methods only ours consistently maintains the required ground constraint of $2\pm0.2m$ for humans while also achieving the highest average distance from obstacles.}
\vspace{-3ex}
\begin{table}
\setlength{\belowcaptionskip}{0pt}
\centering
\begin{tabular}{|c|c|c|c|}
\hline
Dataset & Algorithm & Avg. dis. to ground (m) & Avg. dis. to obstacle (m)\\
\hline

\multirow{4}{*}{\textit{Office 1}}& PRM & 1.71 & 0.286 \\
\cline{2-4}
& A*& 1.72 & 0.187 \\
\cline{2-4}
& CHOMP + 1 GPDF & 2.38 & 0.2 \\
\cline{2-4}
& \textbf{Ours} & \textbf{1.98} & \textbf{0.315} \\
\hline

\multirow{4}{*}{\textit{Cave circuit}}& PRM & 1.68 & 9.13\\
\cline{2-4}
& A*& 4.7 & 8.63  \\
\cline{2-4}
& CHOMP + 1 GPDF & 4.7 & 8.63  \\
\cline{2-4}
& \textbf{Ours} & \textbf{2.012} & \textbf{9.48} \\
\hline

\end{tabular}
\caption{\Rev{Average distance from ground and obstacles given a human on uneven terrains}}
\vspace{-10ex}
\label{tab:avg_distances}
\end{table}
\\


     
     

 \subsection{Discussion and Limitations}

The results show the superior performance of our approach in cluttered spaces. Our formulation is able to tackle uneven terrains as well. Moreover, our approach can easily be adapted to any agent, robot or human, and can find a safe and feasible path accordingly.
The results show that our approach leads to trajectories with high success rates and is always feasible for a ground-based system. 
Furthermore, since we opt for a trajectory optimisation approach, the weights might need to be tuned for other datasets to achieve consistent performances and results. Finally, we assume a cylindrical model for all the agents. However, there are ground-based systems that are not cylindrical, which are not taken into account in our current formulation. 

\section{Conclusion}
This work proposed a method for safe navigation that captures free space information with a multiresolution quadtree and two Gaussian Process distance fields. The sparse, free-space graph built from the quadtree successfully extracts a good approximation of a trajectory, which is then used to optimise and constrict the motion. 
The proposed GPDFs are used to find a 3D trajectory that is safe, smooth and feasible for ground-constrained mobile systems. The Ground GPDF serves two purposes: 1) to enable the obstacle points classification and 2) to maintain the height of the trajectory constant. The Obstacle GPDF is efficiently computed from the 2D projection and used in the trajectory optimisation in the collision loss. Results on 3D real-world datasets show better performance when compared to the traditional CHOMP using one GPDF, A* and PRM. Future work will look into incremental mapping and safe planning for dynamic scenes and introduce a segmentation network to classify ground points online, including elevated surfaces. 

\bibliographystyle{styles/bibtex/splncs03.bst}

\bibliography{root}

%
%







\end{document}